# Trochoid Search Optimization

Abdesslem Layeb

Computer science and its application department, NTIC faculty, university of Constantine 2

Abdesslem.layeb@univ-constantine2.dz

**Abstract :**

The Trochoid Optimization Algorithm (TSO) presents a novel metaheuristic approach to addressing optimization challenges across diverse domains. Rooted in trochoid motion principles, TSO adeptly balances exploration and exploitation to efficiently navigate complex search spaces and uncover high-quality solutions. This paper provides a comprehensive overview of TSO, elucidating its key components, operational mechanisms, and strategic phases, including initialization, exploration, exploitation, and deep intensification search. Comparative analyses against state-of-the-art algorithms across various benchmark functions showcase TSO's competitiveness, demonstrating its ability to achieve competitive mean values, low standard deviations, and explore diverse solution spaces effectively. Moreover, discussions on future research directions underscore TSO's potential for further refinement and application in real-world scenarios. Overall, TSO presents a promising avenue for addressing complex optimization challenges and warrants continued exploration and development in the field of metaheuristic optimization.

**The code link: https://www.mathworks.com/matlabcentral/fileexchange/156677-trochoid-search-optimization**

1. Introduction

Optimization plays a pivotal role in numerous fields, seeking to maximize or minimize key quantities known as optimization problems. For decades, researchers have developed diverse optimization techniques, broadly classified as deterministic and stochastic [1, 2]. Deterministic methods, though predictable, can become trapped in local minima when optimizing complex landscapes. To address this, stochastic optimization techniques offer probabilistic approaches, often achieving superior results within practical timeframes [2].

Nature-inspired algorithms have gained prominence for tackling challenging problems, drawing inspiration from biological, physical, ethological, and swarm-based phenomena [3]. Metaheuristics excel due to their flexibility, avoidance of local optima, and ease of implementation [4]. These algorithms fall into two main groups: single-based and population-based approaches [5], with population-based methods demonstrating superior search space exploration and global optimum discovery. Let's explore the four primary classes of population-based metaheuristics:

- **Swarm Intelligence (SI)** algorithms cleverly mimic collective behaviors found within nature. Examples include the Whale Optimization Algorithm (WOA) [6], Particle Swarm Optimization (PSO) [7], and many others like the Grey Wolf Optimizer (GWO) [8] and Ant Colony Optimization (ACO) [9].
- **Physics-based algorithms** ingeniously leverage the fundamental laws of physics. Notable examples include Simulated Annealing (SA) [10], Gravitational Search Algorithm (GSA) [11], Charged System Search (CSS) [12], and a multitude of others inspired by electromagnetism, water cycles, and even light spectra.
- **Evolutionary algorithms (EAs)** elegantly simulate biological evolution through mechanisms like mutation and natural selection. Genetic Algorithms (GA) [13] pioneered this field, paving the way for techniques such as Differential Evolution (DE) [14], Biogeography-Based Optimizer (BBO) [15], and others.
- **Human-based algorithms** draw inspiration from human social dynamics and behaviors. These include Teaching-Learning Based Optimization (TLBO) [16], Exchange Market Algorithm (EMA) [17], and even algorithms mirroring political processes or sporting competitions.
- **Mathematical algorithms** use mathematic concepts to make optimization. Tangent Search Algorithm (TSA): Drawing inspiration from the mathematical tangent function, TSA employs a model grounded in this function to steer solutions towards improved outcomes. It strikes a balance between exploration and exploitation within the search space and incorporates an escape mechanism to circumvent local minima [18]. Geometric Mean Optimizer (GMO) harnesses the power of the geometric mean to concurrently assess the fitness and diversity of search agents. By assigning the geometric mean of the scaled objective values of an agent's opposites as its weight, GMO offers guidance in navigating optimization challenges [19]. Spheric Search likely revolves around the notion of traversing the search space in a spherical manner, ensuring a thorough and uniform distribution of exploration. It might be associated with the Spherical Search Optimizer (SSO), known for its spherical search strategy in optimization [20]. The Sine Cosine algorithm (SCA) is inspired by the periodic nature of sine and cosine functions. Their oscillation between -1 and 1 allows for balanced exploration (searching broadly) and exploitation (refining good solutions) throughout the optimization process [21].

Despite the rich toolbox of metaheuristics, the No Free Lunch theorem [22] reminds us there's no single algorithm that reigns supreme across all problem domains. This fuels the ongoing pursuit of novel, high-performance global optimization approaches. This paper introduces the Trochoid Optimization Algorithm (TSO), drawing inspiration from the trochoid motion of a point.

TSO, designed to strike a balance between exploration and intensification, comprises two sequential phases: a population search phase for exploration and a subsequent local search phase for intensification around the best solution obtained in the first phase. The TSO algorithm boasts simplicity, guided by a single mathematical equation, and employs a variable step-size to control movement magnitude within the optimization process.

The remainder of the paper is structured as follows: Section 2 delineates the TSO algorithm, providing comprehensive details. Section 3 encompasses experimental results and ensuing discussions. Finally, Section 4 concludes the study

## 2. Algorithm description
### 2.1. Algorithm inspiration

Trochoid Search Optimization (TSO) algorithm is inspired from the trochoidal motion of a point. Geometrically, a trochoid is a curve obtained by plotting the movement described by a point linked to a rolling disc (without sliding) on a line. This term is proposed by the mathematician Roberval (1602-1675) who adapted it from ancient Greek τροχοειδής, trokhoeidês ("circular, round").

Let a disc of radius R rolling without sliding on a line L, the center C moves parallel to L, and all the other points P in the plane attached to the circle form a set of points called the trochoid, let CP = B. Depending on whether P is in the disc (B <1), or on its circumference (B = 1), or outside (B> 1), the trochoid has different curves (figure. 1). The parametric equations of a trochoid on the x-axis y-axis, are [23]:

$$x = R * (\theta + B * \sin(\theta))$$
$$y = R(1 - B * \cos(\theta))$$

Eq.1

with θ is the angle variable describing the rotation of the circle.

For B < 1, the curve is also called curtate cycloid and looks like a sinusoid, and it is one if the term is neglected in x.

For B = 1, we get the cycloid.

For B > 1, the curve is also called prolate cycloid and can assume various shapes, with more and more double points as d increases.

The diverse motions of the trochoid, governed by its mathematical equation, offer a fertile ground for optimization. The TSO algorithm adeptly transitions between exploration (wide search) and intensification (localized refinement), rendering it a promising tool for addressing optimization problems (see Figure 1).

Eq.1

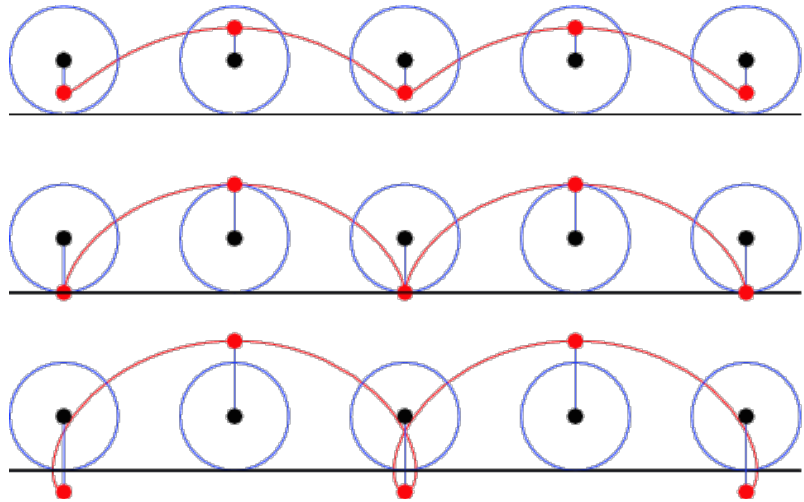

**Figure1.** The different types of trochoids (Weisstein, 2021)

**2.2. Modelling the algorithm**

The Trochoid Optimization Algorithm (TSO) integrates two distinct search strategies—global exploration and local exploitation—guided by the fundamental equations of the trochoid curve. TSO orchestrates the movement of current solutions (points) towards novel ones by leveraging these strategies. The algorithm initiates by creating a population randomly, following which one of the search strategies is activated based on a designated parameter, denoted as Pswich. Additionally, an optional deep intensification procedure can be employed to augment the effectiveness of the STO algorithm (figure 2) in the last iteration of the algorithm.

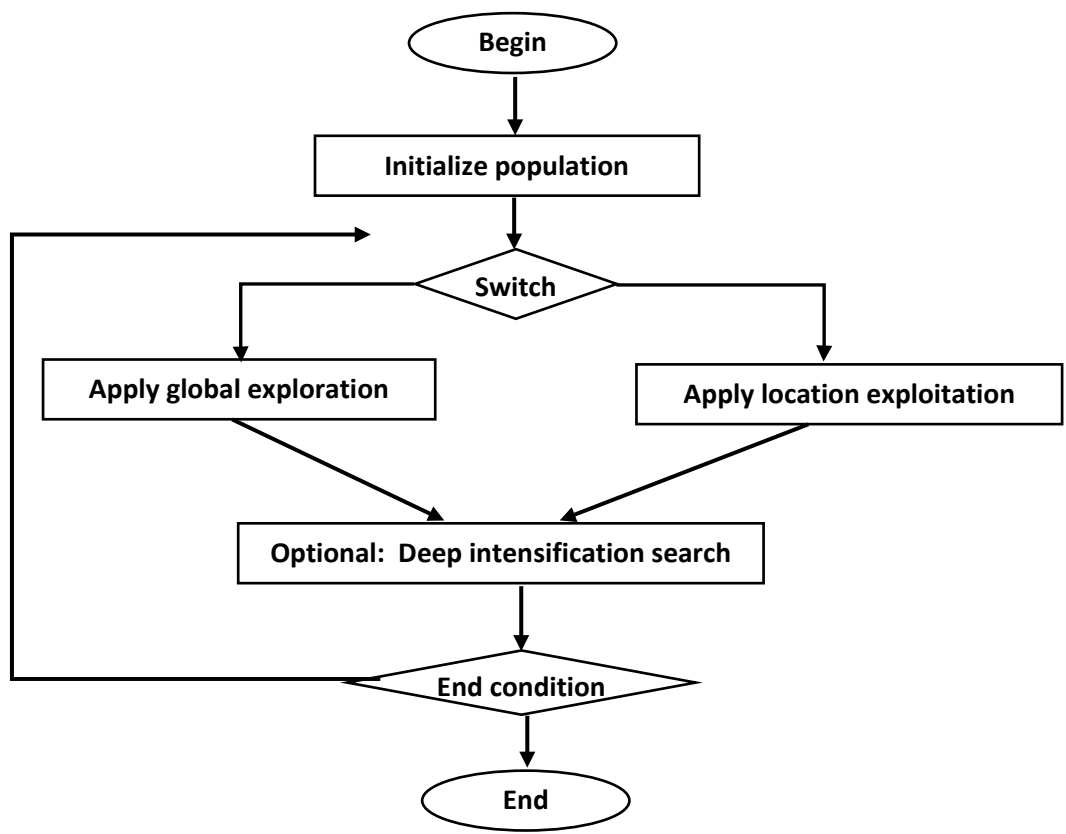

**Figure 2.** The flowchart of TSO

At its core, TSO embodies a hybrid nature, combining elements of global and local search techniques to achieve a balance between diversification and intensification. Through a series of perturbations and adaptive search mechanisms, TSO explores the solution space while concurrently exploiting promising regions to refine the solutions iteratively. This dynamic equilibrium between exploration and exploitation enables TSO to efficiently navigate complex, high-dimensional search spaces, making it suitable for a wide range of optimization tasks.

The salient features of the Trochoid Search Algorithm include:

1. **Adaptive Perturbation**: TSO employs perturbation strategies to diversify the search process, with the perturbation rate dynamically adjusted to balance exploration and exploitation.

2. **Intensification and Exploration**: By incorporating both intensification and exploration phases, TSO strikes a harmonious balance between exploiting known promising regions and exploring new solution spaces.

3. **Dynamic Radius Control**: The algorithm dynamically adjusts the search radius based on the current iteration count, ensuring effective exploration during the early stages and focusing on exploitation in the later stages.

*2.2.1 Initialization of the initial population*
Similar to other population-based optimization algorithms, the Trochoid Search Algorithm (TSO) initiates its exploration by generating a random initial population within the bounds of the solution space. This phase lays the foundation for the subsequent iterative search process, providing an initial set of candidate solutions distributed across the problem space.

X0= lb+(ub-lb).*rand(D);

Where lb, ub are the lower and upper bounds of the problem, rand function generates uniformly distributed random numbers in the range [0, 1], and D is the dimension of the problem. This formulation ensures that X0 exhibits a uniform distribution across the search space defined by the problem's boundaries. The uniform distribution of the initial population enables TSA to effectively explore the entire solution space, laying the groundwork for its adaptive search strategies to navigate complex optimization landscapes. As the algorithm progresses, these initial solutions serve as the starting point for the iterative exploration and exploitation phases, driving the convergence towards optimal solutions

*2.2.2 Exploration and exploitation of the search space*
Optimization algorithms hinge on the delicate balance between exploring and exploiting the search space. Augmenting exploration enhances diversity and avoids entrapment within local minima; however, an excessive focus on exploration can delay optimization progress, potentially leading to algorithmic divergence. Conversely, an emphasis on exploitation accelerates the search process, albeit predominantly converging towards local rather than global minima. In either case, optimization algorithms, whether gradient-based or gradient-free, utilize search directions to guide solutions towards improved states.

During the iterative process, the Trochoid Optimization Algorithm (TSO) dynamically explores and adapts its solutions to navigate the search space effectively. Each iteration, controlled by the termination criterion itrmax, involves traversing the population of candidate solutions. For each individual solution x in the population, a random solution xr is selected to facilitate exploration.
Within a nested loop over the dimensions of the solution space, TSO evaluates whether to perform exploration or exploitation based on a random probability threshold *Pswitch*. If exploitation is chosen, a random dimension k is selected, and a trochoid motion is orchestrated around the current best solution (optX) to modify the selected dimensions. The radius of the trochoid motion is dynamically adjusted based on the current iteration count (itr) and the total number of iterations (itrmax). The exploitation search is governed by the following equation:

$$X_i^j = optx_i^j + R * (1 - B * \sin(\theta))$$
$$X_i^k = optx_i^k + R * (\theta - B * \cos(\theta))$$

Here, $X_i^j$ and $X_i^k$ represent the modified components of the solution vector X corresponding to dimensions j and k, respectively. $optx_i^j$ and $optx_i^k$ denote the values of the current best solution optX in dimensions j and k respectively. $R$ signifies the radius of the trochoid motion, dynamically adjusted based on the current iteration count and the total number of iterations. θ represents a random angle, and $B$ is a random factor.
In these equations, the trochoid motion around the current best solution optX guides the modification of solution components j and k. The adjustment in each dimension is influenced by both the angle θ and the random factor $B$, ensuring a controlled and strategic exploration of the solution space. By leveraging these equations, TSO orchestrates a balance between global and local search strategies, facilitating effective exploitation of promising solutions while maintaining diversity and adaptability throughout the optimization process.

However, the exploration search is governed by the following equation:

$$X_i^j = X_r^j + R * (1 - B * \sin(\theta))$$
$$X_i^k = X_r^k + R * (\theta - B * \cos(\theta))$$

Here, $X_i^j$ and $X_i^k$ represent the modified components of the solution vector X corresponding to dimensions j and k, respectively. $X_r^j$ and $X_r^k$ denote the values of a randomly selected solution Xr in dimensions j and k. $R$ signifies the radius of the trochoid motion, dynamically adjusted based on the current iteration count and the total number of iterations. $\theta$ represents a random angle, and $B$ is a random factor.

In these equations, the trochoid motion around the randomly selected solution Xr guides the modification of solution components j and k. The adjustment in each dimension is influenced by both the angle $\theta$ and the random factor $B$, ensuring a controlled and strategic exploration of the solution space. By leveraging these equations, TSO fosters diversity and exploration, facilitating the discovery of new regions in the solution space while maintaining adaptability throughout the optimization process.

In the following, we explain in more detail the main parameters of TSO.

*2.2.3 The computation of the TSO parameters*
The important parameters in the above equations is the radius of the trochoid R, B and the angle $\theta$. TSO implements a variable step-size to facilitate convergence towards the optimal solution while ensuring precision. Initially, TSO adopts a large step-size; gradually, as the search progresses, the step-size nonlinearly decreases from one iteration to the next. This adaptive behavior of the step-size aids TSO in achieving an optimal balance between exploration and exploitation.

In the exploitation search phase of the Trochoid Optimization Algorithm (TSO), the computation of the radius $R$, the random factor $B$, and the angle $\theta$ is as follows. The radius $R$ is calculated using the following equation:

$$R = (1 - (\frac{itr}{itrmax}))^{2 \times (\frac{itr}{itrmax})} * d$$

Here, *itr* represents the current iteration count, *itrmax* denotes the maximum number of iterations, d is the search direction. This formula ensures that the radius decreases as the number of iterations increases, guiding the trochoid motion towards convergence.

The direction d is determined by the difference between the current best solution *optX* and the current solution *x*:

$$d = (optX(i) - x(i))$$

This direction guides the trochoid motion towards or away from the current best solution, influencing the exploration process.

The random factor $B$ is generated uniformly in the range [0, 1]. This random factor introduces stochasticity into the exploration process, contributing to the diversity of the search. The angle $\theta$ is generated randomly in the range [0, $\pi$] as follow:

$$\theta = rand \times \pi$$

This random angle determines the direction of the trochoid motion, influencing the exploration of the solution space.

Together, these computations govern the dynamics of the exploitation search in TSO, facilitating a balance between exploration and exploitation as the algorithm iteratively refines and adapts its solutions.

On the other hand, in the exploration phase of the TSO, the computation of the radius $R$, and the random factor B is as follow:

The radius $R$ is calculated using the following equation:

$$R = 15 * sign(1 - 2 \times rand) * log(itr + 10)^2 * \boldsymbol{d}$$

Here, *rand* generates a random number in the range [0, 1]. The sign function ensures that the radius direction is randomly selected, enhancing exploration. The logarithmic term $log(itr + 10)^2$ ensures that the radius decreases with the progression of iterations, promoting convergence towards promising regions of the search space. The direction d is similar to previous one. Similar to exploitation phase, the random factor $B$ is generated randomly in the rage of [0,1].

These adaptive parameters enable the algorithm to effectively explore diverse regions of the solution space while maintaining adaptability and exploration capability throughout the optimization process.

The main steps of the TSO algorithm are given by the following figure.

```
% population creation
 pop =  create population
% population evaluation
 fitness_pop=ones(1,pop_size)*1e30;
% create a best solution
 optX=ub+rand(1,dim)*(ub-lb);
 opt=fun(optX);
 itr=1; % iteration

    while itr<itrmax      % iterations starts

       for j=1: pop_size
         x1=pop(j,:);

         for i=1:dim
           if rand <=Pswitch
             apply global exploration search
           else
             apply local exploitation search
           end

       end
    end

       if rand <=Pescape
          apply escape local procedure

       end

       repair the new solution

       evaluate the new solution

       update the population

    end
```

**Figure 3.** The pseudo code the TSO algorithm

### 2.2.4 Deep intensification search phase

In the deep intensification search phase of the Trochoid Optimization Algorithm (TSO), adjustments to the solution are made when the iteration count exceeds a certain threshold, specifically when $itr \geq 0.8 \times itrmax$.

During this phase, the current solution X is subjected to adjustments aimed at intensifying the search towards promising regions of the solution space. These adjustments are governed by the following process:

1. If a randomly generated number is less than or equal to 0.2, a tangent function of a randomly generated angle is applied to the solution, leading to adjustments along a direction perpendicular to the solution space boundaries. This introduces a level of randomness into the search process, promoting exploration in unexplored regions.

2. Alternatively, if the randomly generated number exceeds 0.2, a step-wise adjustment is performed. This adjustment is determined by a combination of a random factor and the logarithmically-scaled iteration count. Specifically, the solution X is adjusted based on the magnitude and direction of the difference between x and a randomly weighted difference between the current best solution *optX* and X. This process facilitates a more directed search towards potentially optimal solutions while still allowing for exploration.

Together, these adjustments in the deep intensification search phase aim to further refine and intensify the search process, enabling TSO to converge towards optimal or near-optimal solutions, particularly in the later stages of the optimization process.

```
       if rand <=0.5 && itr <=0.5*itrmax
         X = X+ tan(rand*pi).*(ub-lb);
       else
          step = 15*(1-2*rand)/log(1+itr);
          X= X+ step.*(X-rand*( optX -X));
       end
     end
```

**Figure 4.** The pseudo code of the escape local search

## 3. Experimental settings

The effectiveness of this new approach, referred to as TSO, was evaluated using a benchmark provided by the CEC2022 [24], which includes a range of challenging unimodal, multimodal, and hybrid functions (as listed in Table 1). It is important to note that all functions were shifted to avoid any bias towards a particular algorithm. The experimental setup involved using an Intel Core i3 processor with 4 GB RAM, running Windows 10 (64 bit) and MATLAB 2017a. To measure the performance of TSO, we compared it against well-known optimizers such as TSA[18], LSHADE[25], APGSK_IMODE [26], EO[27], MDEW[5], and LBPSO[28]. The parameters used for TSO and the other algorithms are summarized in the table 2.

The population size is fixed to 60, and the number of maximum iterations is 10000*D where D=30. The experimental findings, available in Tables 1 and 2, highlight the performance of the TSO algorithm. It's evident from the results that the TSO algorithm showcases notable performance across the test functions. Additionally, the utilization of the escape local procedure has demonstrated significant benefits in enhancing the algorithm's efficacy.

**Table 1.** Summary of the CEC'22 Test Suite Search, range: [-100,100]

|  | No. | Functions | opt* |
|---|---|---|---|
| Unimodal Function | 1 | Shifted and full Rotated Zakharov Function | 300 |
| Basic Functions | 2 | Shifted and full Rotated Rosenbrock's Function | 400 |
|  | 3 | Shifted and full Rotated Expanded Schaffer's $f6$ Function | 600 |
|  | 4 | Shifted and full Rotated Non-Continuous Rastrigin's Function | 800 |
|  | 5 | Shifted and full Rotated Levy Function | 900 |
| Hybrid Functions | 6 | Hybrid Function 1 ($N = 3$) | 1800 |
|  | 7 | Hybrid Function 2 ($N = 6$) | 2000 |
|  | 8 | Hybrid Function 3 ($N = 5$) | 2200 |
| Composition Functions | 9 | Composition Function 1 ($N = 5$) | 2300 |
|  | 10 | Composition Function 2 ($N = 4$) | 2400 |
|  | 11 | Composition Function 3 ($N = 5$) | 2600 |
|  | 12 | Composition Function 4 ($N = 6$) | 2700 |

Table 2. parameters setting.

| Name | References | Parameters |
|---|---|---|
| TSO |  | pop_size =60, perturbation rate  pm=0.1; Switch parameter pr_intens=0.5; |
| MDEW | [5] | F1=0.25, F2= $rand(-,+) * 0.5$, CR=0.75, popsize=50 |
| TSA | [18] | *Pswitch=0.3, Pesc=0.8, popsize=50* |
| LSHADE | [25] | p_best_rate = 0.11; arc_rate = 1.4; pop_size = 18* D; memory_size = 5; |

| | | | | | | | | | | | | |
|---|---|---|---|---|---|---|---|---|---|---|---|---|
| EO | | [27] | a1=2, a2=1, GP=0.5, *popsize=50* | | | | | | | | | |
| | | | max_pop_size = pop_size; | | | | | | | | | |
| | | | min_pop_size = 4; | | | | | | | | | |
| APGSK_IMODE | | [26] | F =0.50, CR=0.9, | | | | | | | | | |
| | | | p_best_rate = 0.5; | | | | | | | | | |
| | | | p_best_rate_min = 0.15; | | | | | | | | | |
| | | | memory_size = 5; | | | | | | | | | |
| | | | pop_size = 30* D; | | | | | | | | | |
| | | | max_pop_size = pop_size; | | | | | | | | | |
| | | | min_pop_size = 4; | | | | | | | | | |
| LBPSO | | [28] | iwt = 0.9 - (1 : maxGEN) * (0.7 / maxGEN); | | | | | | | | | |
| | | | c = 1.49445; | | | | | | | | | |
| | | | I = 1; % max immigration rate | | | | | | | | | |
| | | | E = 1; % max emigration rate | | | | | | | | | |
| | | | MigrateModel = 5; | | | | | | | | | |
| | | | pop_size =40 | | | | | | | | | |

The results presented in Table 3 show the comparative performance of the Trochoid Optimization Algorithm (TSO) against other optimization algorithms across various 20-dimensional CEC2022 test functions. The table provides statistical measures such as mean, standard deviation (std), minimum (min), and maximum (max) values of the objective function evaluations obtained by each algorithm across different test functions. TSO demonstrates competitive performance across the evaluated metrics compared to other algorithms such as TSA, LSHADE, EO, APGSK_IMODE, LBPSO, and MDEW. The performance of each algorithm varies across different test functions, indicating the importance of algorithm selection based on problem characteristics.

**Table 3.** Comparative results of TSO, and other algorithms on 20D CEC2022 test functions

| | | cec22_01 | cec22_02 | cec22_03 | cec22_04 | cec22_05 | cec22_06 | cec22_07 | cec22_08 | cec22_09 | cec22_10 | cec22_11 | cec22_12 |
|---|---|---|---|---|---|---|---|---|---|---|---|---|---|
| TSA | mean | 0.000E+00 | 3.747E+01 | 3.831E-02 | 3.414E+01 | 3.604E+01 | 3.358E+01 | 2.935E+01 | 2.094E+01 | 1.808E+02 | 1.004E+02 | 2.409E+02 | 2.531E+02 |
| | std | 0.000E+00 | 2.078E+01 | 8.237E-02 | 7.636E+00 | 4.866E+01 | 1.702E+01 | 7.254E+00 | 2.357E+00 | 0.000E+00 | 8.256E-02 | 1.559E+02 | 6.300E+00 |
| | min | 0.000E+00 | 8.327E-04 | 0.000E+00 | 1.990E+01 | 2.086E+00 | 6.353E+00 | 1.285E+01 | 8.912E+00 | 1.808E+02 | 1.003E+02 | 0.000E+00 | 2.378E+02 |
| | max | 0.000E+00 | 4.908E+01 | 3.423E-01 | 4.900E+01 | 2.333E+02 | 6.739E+01 | 4.894E+01 | 2.312E+01 | 1.808E+02 | 1.006E+02 | 4.000E+02 | 2.656E+02 |
| LSHADE | mean | 0.00E+00 | 3.75E+01 | 3.83E-02 | 3.41E+01 | 3.60E+01 | 3.36E+01 | 2.94E+01 | 2.09E+01 | 1.81E+02 | 1.00E+02 | 2.41E+02 | 2.53E+02 |
| | std | 0.00E+00 | 2.08E+01 | 8.24E-02 | 7.64E+00 | 4.87E+01 | 1.70E+01 | 7.25E+00 | 2.36E+00 | 0.00E+00 | 8.26E-02 | 1.56E+02 | 6.30E+00 |
| | min | 0.00E+00 | 8.33E-04 | 0.00E+00 | 1.99E+01 | 2.09E+00 | 6.35E+00 | 1.29E+01 | 8.91E+00 | 1.81E+02 | 1.00E+02 | 0.00E+00 | 2.38E+02 |
| | max | 0.00E+00 | 4.91E+01 | 3.42E-01 | 4.90E+01 | 2.33E+02 | 6.74E+01 | 4.89E+01 | 2.31E+01 | 1.81E+02 | 1.01E+02 | 4.00E+02 | 2.66E+02 |
| EO | mean | 0.00E+00 | 4.89E+01 | 0.00E+00 | 4.69E+00 | 0.00E+00 | 7.20E-01 | 1.35E+01 | 2.03E+01 | 1.81E+02 | 1.00E+02 | 3.03E+02 | 2.33E+02 |
| | std | 0.00E+00 | 7.65E-01 | 0.00E+00 | 1.17E+00 | 0.00E+00 | 5.66E-01 | 6.53E+00 | 1.58E+00 | 0.00E+00 | 3.22E-02 | 1.83E+01 | 1.35E+00 |
| | min | 0.00E+00 | 4.49E+01 | 0.00E+00 | 1.99E+00 | 0.00E+00 | 1.57E-01 | 3.96E+00 | 1.34E+01 | 1.81E+02 | 1.00E+02 | 3.00E+02 | 2.32E+02 |
| | max | 0.00E+00 | 4.91E+01 | 0.00E+00 | 5.98E+00 | 0.00E+00 | 2.31E+00 | 2.19E+01 | 2.12E+01 | 1.81E+02 | 1.00E+02 | 4.00E+02 | 2.38E+02 |
| APGSK_IMODE | mean | 4.26E-05 | 5.10E+01 | 3.93E-02 | 3.96E+01 | 1.96E+00 | 2.62E+03 | 3.91E+01 | 2.71E+01 | 1.81E+02 | 6.36E+02 | 2.97E+02 | 2.48E+02 |
| | std | 8.25E-05 | 1.23E+01 | 1.63E-01 | 1.10E+01 | 2.77E+00 | 4.16E+03 | 1.44E+01 | 2.24E+01 | 2.24E-04 | 5.68E+02 | 6.15E+01 | 5.87E+00 |
| | min | 0.00E+00 | 2.94E+00 | 2.86E-06 | 2.29E+01 | 0.00E+00 | 9.48E+01 | 2.23E+01 | 2.14E+01 | 1.81E+02 | 1.00E+02 | 0.00E+00 | 2.39E+02 |
| | max | 3.12E-04 | 7.08E+01 | 8.85E-01 | 5.67E+01 | 1.26E+01 | 1.72E+04 | 7.03E+01 | 1.45E+02 | 1.81E+02 | 1.86E+03 | 4.00E+02 | 2.59E+02 |
| LBPSO | mean | 1.53E+01 | 4.91E+01 | 6.70E-07 | 2.88E+01 | 2.95E-01 | 6.32E+01 | 2.70E+01 | 2.25E+01 | 1.81E+02 | 1.01E+02 | 3.00E+02 | 2.38E+02 |
| | std | 3.63E+01 | 2.50E-07 | 6.10E-07 | 5.13E+00 | 3.92E-01 | 2.31E+01 | 3.20E+00 | 1.36E+00 | 0.00E+00 | 5.22E-02 | 0.00E+00 | 2.71E+00 |
| | min | 9.68E-03 | 4.91E+01 | 4.00E-08 | 1.98E+01 | 0.00E+00 | 2.89E+01 | 2.27E+01 | 1.71E+01 | 1.81E+02 | 1.00E+02 | 3.00E+02 | 2.32E+02 |
| | max | 1.62E+02 | 4.91E+01 | 2.27E-06 | 3.75E+01 | 1.68E+00 | 1.07E+02 | 3.34E+01 | 2.45E+01 | 1.81E+02 | 1.01E+02 | 3.00E+02 | 2.43E+02 |
| TSO | mean | 2.57E+01 | 5.08E+01 | 1.33E-08 | 5.04E+01 | 2.47E-02 | 3.71E+03 | 1.94E+01 | 2.33E+01 | 1.81E+02 | 9.14E+01 | 3.03E+02 | 2.41E+02 |
| | std | 2.95E+01 | 5.33E+00 | 5.95E-08 | 1.25E+01 | 7.43E-02 | 3.31E+03 | 5.52E+00 | 9.61E-01 | 1.59E-03 | 4.17E+01 | 6.69E+01 | 5.39E+00 |
| | min | 2.38E-01 | 4.91E+01 | 0.00E+00 | 2.75E+01 | 3.41E-13 | 5.91E+01 | 4.76E+00 | 2.06E+01 | 1.81E+02 | 3.12E-01 | 1.82E-12 | 2.33E+02 |
| | max | 1.22E+02 | 6.93E+01 | 3.19E-07 | 7.91E+01 | 3.83E-01 | 1.49E+04 | 3.03E+01 | 2.46E+01 | 1.81E+02 | 2.20E+02 | 4.00E+02 | 2.52E+02 |
| MDEW | mean | 2.57E+01 | 5.08E+01 | 1.33E-08 | 5.04E+01 | 2.47E-02 | 3.71E+03 | 1.94E+01 | 2.33E+01 | 1.81E+02 | 9.14E+01 | 3.03E+02 | 2.41E+02 |

|  |  |  |  |  |  |  |  |  |  |  |  |  |
|---|---|---|---|---|---|---|---|---|---|---|---|---|
|  | std | 4.02E-02 | 1.59E+01 | 9.52E-02 | 1.10E+01 | 4.53E+00 | 3.51E+03 | 9.04E+00 | 5.75E-01 | 2.43E-03 | 7.27E+01 | 6.69E+01 | 8.59E+00 |
|  | min | 3.59E-08 | 4.41E+00 | 7.98E-08 | 1.39E+01 | 1.02E-12 | 4.32E+01 | 7.65E+00 | 1.98E+01 | 1.81E+02 | 3.70E+00 | 4.55E-13 | 2.33E+02 |
|  | max | 1.81E-01 | 7.26E+01 | 4.26E-01 | 6.15E+01 | 1.73E+01 | 1.27E+04 | 5.28E+01 | 2.26E+01 | 1.81E+02 | 2.38E+02 | 4.00E+02 | 2.74E+02 |

Table 3 presents a comparative analysis of the Trochoid Optimization Algorithm (TSO) against other algorithms on 10-dimensional CEC2022 test functions. TSO showcases competitive mean values across most test functions, implying its capacity to attain solutions proximate to the global optimum, on average. With relatively low standard deviation values, TSO exhibits consistent performance across various runs, showing resilience against performance fluctuations. Moreover, TSO achieves reasonable minimum and maximum values, underscoring its adeptness in efficiently exploring the search space to uncover both promising and sub-optimal solutions. Notably, TSO demonstrates consistency in performance, evidenced by the similarity in mean, standard deviation, minimum, and maximum values across multiple test functions.

Overall, similar to its performance in the 20-dimensional case, TSO shows promising performance compared to other optimization algorithms across various performance metrics in the 10-dimensional scenario. Further analysis, such as statistical tests or convergence analysis, may provide additional insights into TSO's performance compared to other algorithms.

**Table 4.** Comparative results of TSO and other algorithms on 10D CEC2022 test functions

|  |  | cec22_01 | cec22_02 | cec22_03 | cec22_04 | cec22_05 | cec22_06 | cec22_07 | cec22_08 | cec22_09 | cec22_10 | cec22_11 | cec22_12 |
|---|---|---|---|---|---|---|---|---|---|---|---|---|---|
| TSA | mean | 0.00E+00 | 3.57E-02 | 1.11E-06 | 1.05E+01 | 2.98E-03 | 2.49E+01 | 4.65E-01 | 2.41E+00 | 2.29E+02 | 1.08E+01 | 0.00E+00 | 1.63E+02 |
|  | std | 0.00E+00 | 4.61E-02 | 2.32E-06 | 4.22E+00 | 1.64E-02 | 2.09E+01 | 5.05E-01 | 6.04E+00 | 0.00E+00 | 2.49E+01 | 0.00E+00 | 1.77E+00 |
|  | min | 0.00E+00 | 6.90E-07 | 0.00E+00 | 3.98E+00 | 0.00E+00 | 4.39E-01 | 1.00E-08 | 1.41E-03 | 2.29E+02 | 1.25E-01 | 0.00E+00 | 1.59E+02 |
|  | max | 0.00E+00 | 1.93E-01 | 1.13E-05 | 2.39E+01 | 8.95E-02 | 7.27E+01 | 9.96E-01 | 2.03E+01 | 2.29E+02 | 1.00E+02 | 0.00E+00 | 1.65E+02 |
| LSHADE | mean | 0.00E+00 | 5.47E+00 | 0.00E+00 | 2.72E+00 | 0.00E+00 | 2.57E-01 | 5.95E-03 | 2.18E+00 | 2.29E+02 | 1.00E+02 | 0.00E+00 | 1.61E+02 |
|  | std | 0.00E+00 | 2.30E+00 | 0.00E+00 | 8.64E-01 | 0.00E+00 | 1.39E-01 | 1.55E-02 | 2.72E+00 | 0.00E+00 | 1.83E-02 | 0.00E+00 | 1.61E+00 |
|  | min | 0.00E+00 | 3.99E+00 | 0.00E+00 | 3.31E-04 | 0.00E+00 | 3.94E-02 | 0.00E+00 | 1.48E-01 | 2.29E+02 | 1.00E+02 | 0.00E+00 | 1.59E+02 |
|  | max | 0.00E+00 | 8.92E+00 | 0.00E+00 | 3.98E+00 | 0.00E+00 | 4.98E-01 | 8.40E-02 | 1.04E+01 | 2.29E+02 | 1.00E+02 | 0.00E+00 | 1.65E+02 |
| EO | mean | 0.00E+00 | 8.79E+00 | 3.52E-05 | 1.13E+01 | 8.76E-02 | 2.68E+03 | 1.87E+01 | 1.88E+01 | 2.29E+02 | 1.52E+02 | 9.00E+01 | 1.64E+02 |
|  | std | 0.00E+00 | 1.30E+01 | 9.85E-05 | 4.90E+00 | 1.63E-01 | 2.30E+03 | 7.02E+00 | 6.93E+00 | 0.00E+00 | 5.65E+01 | 1.49E+02 | 1.37E+00 |
|  | min | 0.00E+00 | 1.75E-03 | 0.00E+00 | 2.99E+00 | 0.00E+00 | 2.48E+01 | 9.10E-07 | 8.89E-01 | 2.29E+02 | 1.00E+02 | 0.00E+00 | 1.60E+02 |
|  | max | 0.00E+00 | 7.57E+01 | 3.73E-04 | 2.67E+01 | 5.44E-01 | 6.42E+03 | 2.90E+01 | 2.55E+01 | 2.29E+02 | 2.19E+02 | 4.00E+02 | 1.66E+02 |
| APGSK_IMODE | mean | 0.00E+00 | 1.00E-08 | 0.00E+00 | 4.92E+00 | 0.00E+00 | 2.86E-01 | 3.47E-04 | 1.11E+00 | 2.22E+02 | 1.00E+02 | 0.00E+00 | 1.61E+02 |
|  | std | 0.00E+00 | 0.00E+00 | 0.00E+00 | 1.23E+00 | 0.00E+00 | 2.52E-01 | 7.58E-04 | 4.92E-01 | 4.19E+01 | 4.03E-02 | 0.00E+00 | 1.56E+00 |
|  | min | 0.00E+00 | 0.00E+00 | 0.00E+00 | 1.99E+00 | 0.00E+00 | 2.15E-02 | 0.00E+00 | 3.26E-01 | 1.00E+02 | 1.00E+02 | 0.00E+00 | 1.59E+02 |
|  | max | 0.00E+00 | 5.00E-08 | 0.00E+00 | 7.11E+00 | 0.00E+00 | 1.43E+00 | 3.12E-03 | 2.31E+00 | 2.29E+02 | 1.00E+02 | 0.00E+00 | 1.64E+02 |
| LBPSO | mean | 0.00E+00 | 4.77E+00 | 0.00E+00 | 3.69E+00 | 0.00E+00 | 1.60E+02 | 1.58E-01 | 7.08E+00 | 2.29E+02 | 1.07E+02 | 0.00E+00 | 1.65E+02 |
|  | std | 0.00E+00 | 3.59E+00 | 0.00E+00 | 1.17E+00 | 0.00E+00 | 2.71E+02 | 2.98E-01 | 8.20E+00 | 0.00E+00 | 2.67E+01 | 0.00E+00 | 7.87E-01 |
|  | min | 0.00E+00 | 2.43E-02 | 0.00E+00 | 2.07E+00 | 0.00E+00 | 4.57E+00 | 0.00E+00 | 3.38E-02 | 2.29E+02 | 1.00E+02 | 0.00E+00 | 1.63E+02 |
|  | max | 4.00E-08 | 8.92E+00 | 0.00E+00 | 6.51E+00 | 0.00E+00 | 1.11E+03 | 9.95E-01 | 2.10E+01 | 2.29E+02 | 2.06E+02 | 0.00E+00 | 1.67E+02 |
| TSO | mean | 4.19E-03 | 4.62E+00 | 3.79E-15 | 8.78E+00 | 3.49E-13 | 6.30E+02 | 1.63E+00 | 5.84E+00 | 2.29E+02 | 1.04E+02 | 1.51E+01 | 1.65E+02 |
|  | std | 2.19E-02 | 3.57E+00 | 2.08E-14 | 3.61E+00 | 5.23E-13 | 9.36E+02 | 5.00E+00 | 8.76E+00 | 2.97E-13 | 1.92E+01 | 4.59E+01 | 1.14E+00 |
|  | min | 2.17E-08 | 1.40E-03 | 0.00E+00 | 2.99E+00 | 1.14E-13 | 2.51E+00 | 6.55E-07 | 6.04E-03 | 2.29E+02 | 1.00E+02 | 0.00E+00 | 1.63E+02 |
|  | max | 1.20E-01 | 8.92E+00 | 1.14E-13 | 1.72E+01 | 3.07E-12 | 4.70E+03 | 2.00E+01 | 2.03E+01 | 2.29E+02 | 2.05E+02 | 1.51E+02 | 1.67E+02 |
| MDEW | mean | 4.19E-03 | 4.62E+00 | 3.79E-15 | 8.78E+00 | 3.49E-13 | 6.30E+02 | 1.63E+00 | 5.84E+00 | 2.29E+02 | 1.04E+02 | 1.51E+01 | 1.65E+02 |
|  | std | 0.00E+00 | 3.56E-04 | 0.00E+00 | 3.34E+00 | 0.00E+00 | 1.57E-01 | 4.88E-01 | 3.73E+00 | 4.13E+01 | 5.28E+00 | 0.00E+00 | 1.07E+00 |
|  | min | 0.00E+00 | 0.00E+00 | 0.00E+00 | 6.97E+00 | 0.00E+00 | 8.93E-02 | 0.00E+00 | 3.22E-02 | 3.00E+00 | 1.27E-01 | 0.00E+00 | 1.61E+02 |
|  | max | 0.00E+00 | 1.93E-03 | 0.00E+00 | 1.99E+01 | 0.00E+00 | 7.86E-01 | 1.00E+00 | 2.03E+01 | 2.29E+02 | 1.69E+01 | 0.00E+00 | 1.65E+02 |

4. **Conclusion**

In conclusion, this paper has introduced the Trochoid Optimization Algorithm (TSO) as a novel metaheuristic approach for optimization problems. Through a comprehensive exploration of its key components and operational mechanisms, we have elucidated how TSO strategically integrates trochoid motion principles into its search process to balance exploration and exploitation effectively. The initialization of the initial population, exploration of the search space, and deep intensification search phases have been meticulously described, highlighting the algorithm's robustness and adaptability.

Furthermore, comparative results presented in Tables 3 and 4 underscore TSO's competitiveness against other state-of-the-art algorithms across various benchmark functions. TSO consistently demonstrates strong performance, achieving competitive mean values, low standard deviations, and reasonable minimum and maximum values across different dimensions and test functions. These results affirm TSO's efficacy in efficiently navigating complex search spaces and uncovering high-quality solutions.

Looking ahead, future research directions could focus on further refining TSO's parameters and mechanisms to enhance its convergence speed and scalability for larger-scale optimization problems. Additionally, investigating TSO's performance on real-world applications and benchmarking against more diverse sets of algorithms would provide valuable insights into its practical utility and generality. Overall, the Trochoid Optimization Algorithm presents a promising avenue for addressing complex optimization challenges across various domains, warranting continued exploration and development in the field of metaheuristic optimization.